\documentclass[10pt,twocolumn,letterpaper]{article}

\usepackage[pagenumbers]{cvpr} % To force page numbers, e.g. for an arXiv version

\usepackage[misc]{ifsym}
\usepackage{multirow}

\definecolor{cvprblue}{rgb}{0.21,0.49,0.74}
\usepackage[pagebackref,breaklinks,colorlinks,allcolors=cvprblue]{hyperref}

\def\paperID{*****} % *** Enter the Paper ID here
\def\confName{CVPR}
\def\confYear{2026}

\title{PointForward V2: Point-Aligned Interaction for Feedforward Driving Reconstruction}

\author{Xianqi Wang\textsuperscript{1, *} \quad Cheng Chi\textsuperscript{2, 4, *} \quad Yuechuan Pu\textsuperscript{2, 4, *} \quad Hongcheng Luo\textsuperscript{2, 4}\\Junda Cheng\textsuperscript{3} \quad Guang Chen\textsuperscript{2, 4} \quad Hangjun Ye\textsuperscript{2, 4} \quad Bing Wang\textsuperscript{2, 4, $\dagger$} \quad Xin Yang\textsuperscript{1, \Letter} \\
[2mm]
\textsuperscript{1}Huazhong University of Science and Technology \quad \textsuperscript{2}Xiaomi EV \quad \textsuperscript{3}HKUST \\ \textsuperscript{4}Beijing Key Laboratory of Intelligent Driving Foundation Models and Simulation\\
}
\begin{document}
\maketitle
\begin{abstract}
Feedforward reconstruction of dynamic driving scenes requires integrating observations across multiple views and time steps into a coherent 3D representation. PointForward established state-of-the-art feedforward driving reconstruction by lifting image observations into sparse 3D queries, demonstrating the effectiveness of point-aligned scene representations. However, these queries are largely processed independently, limiting information exchange across space and time, while direct global attention among tens of thousands of point queries is computationally prohibitive. In this work, we present \textbf{PointForward V2}, which advances point-aligned representation to point-aligned interaction. Our key idea is to mediate query interaction through compact, scene-structured registers. For the static background, frame-wise Static Scene Registers compress point information, communicate across frames, and propagate the aggregated context back to the queries. For each dynamic instance, queries from different frames are aligned in an object-centric canonical space and interact through instance-specific Dynamic Instance Registers. We further employ 4D positional encoding that combines aligned spatial coordinates with original timestamps, enabling spatially aligned yet temporally distinguishable interactions. In this way, PointForward V2 enables scalable spatio-temporal reasoning directly among lifted 3D queries without dense global attention. Experiments on Waymo, nuScenes, and Argoverse 2 demonstrate state-of-the-art performance across large-scale driving benchmarks. Code: \textcolor{magenta}{https://github.com/Windsrain/PointForward-V2}.

\end{abstract}

\footnotetext[0]{
\textsuperscript{*} Equal contribution, 
\textsuperscript{$\dagger$} Project leader,
\textsuperscript{\Letter} Corresponding author.
}

\section{Introduction}
\label{sec:intro}

Accurately reconstructing dynamic driving scenes is fundamental to autonomous driving and supports applications such as novel-view synthesis, simulation, and closed-loop evaluation. Neural Radiance Fields (NeRFs) and 3D Gaussian Splatting (3DGS) have enabled high-fidelity reconstruction, and subsequent methods extend them to large-scale dynamic driving environments through scene decomposition, object-centric modeling, and spatio-temporal representations~\cite{mildenhall2021nerf, kerbl20233d, yan2024street, yang2024emernerf, chen2025omnire}. However, these approaches typically rely on costly per-scene optimization. Recent feedforward methods instead learn cross-scene priors and reconstruct unseen scenes in a single forward pass~\cite{charatan2024pixelsplat, chen2024mvsplat, yang2025storm, chen2026dggt}. PointForward~\cite{chi2026pointforward} further departs from the prevailing pixel-aligned paradigm by lifting image observations into sparse 3D queries and constructing point-aligned scene representations. This formulation achieves state-of-the-art feedforward reconstruction across driving benchmarks, validating point alignment as an effective representation for dynamic driving scenes. Yet the resulting point queries are still processed largely independently: point alignment determines where information is represented, but does not, by itself, enable information exchange between points.

In PointForward~\cite{chi2026pointforward}, each query gathers multi-view evidence by projecting its 3D location back to the input images, sampling the corresponding image features, and aggregating them through spatio-temporal fusion. Consequently, spatially or temporally related queries cannot directly exchange complementary geometric and appearance cues. A natural solution is global self-attention~\cite{vaswani2017attention, dosovitskiy2020image} over the point queries, but its quadratic complexity becomes prohibitive for the tens of thousands of queries required to represent a driving scene. Prior work has explored compact intermediate representations to enable scalable global communication~\cite{lee2019set, jaegle2021perceiver, jaegle2021perceiverio, han2024agent, ngo2025delta, wang2026vggt}. However, directly applying a proxy-token mechanism overlooks the compositional structure of driving scenes, where the static background is shared across frames. Moreover, dynamic instances form independent object-centric components. This raises a question: how can point-aligned queries interact efficiently while respecting the structure of dynamic driving scenes?

\begin{figure*}[t]
    \centering
    \includegraphics[width=1.0\textwidth]{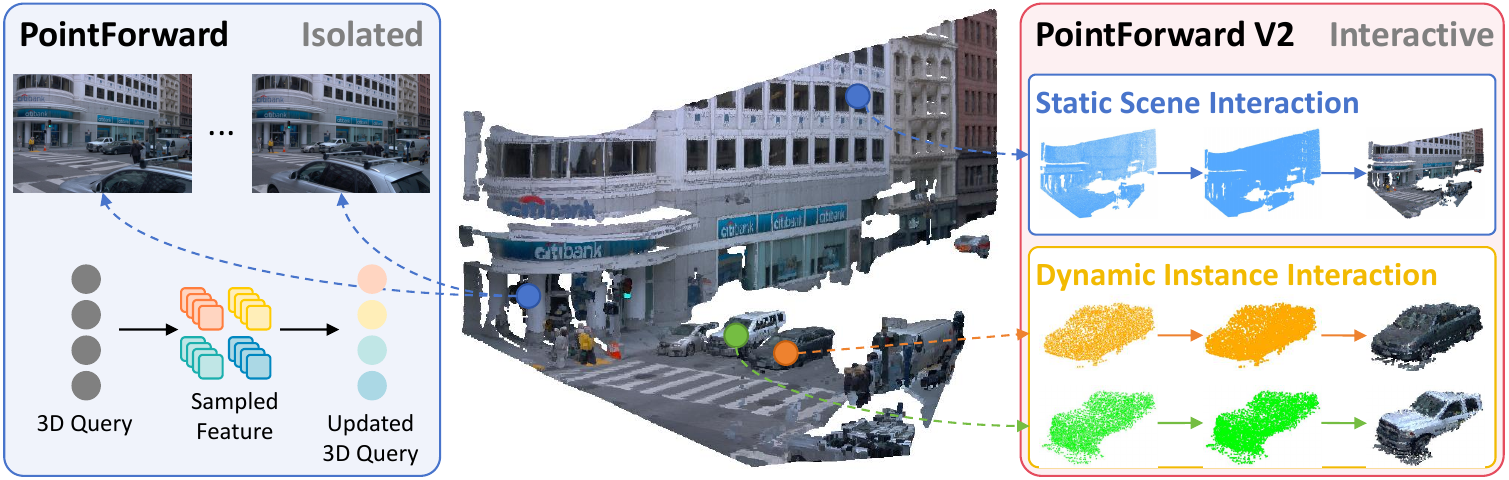}
    \caption{\textbf{From point-aligned representation to interaction.} While PointForward processes 3D queries independently, PointForward V2 enables scene-structured interaction: static queries communicate across frames, while dynamic queries interact within each instance.}
    \label{fig:teaser}
    \vspace{-10px}
\end{figure*}

In this work, we present \textbf{PointForward V2}, a feedforward driving reconstruction framework that advances point-aligned representation to point-aligned interaction. Instead of performing attention directly among point queries, we introduce compact, scene-structured registers as bridges. For the static background, we perform \emph{inter-frame scene interaction} through Static Scene Registers. Point queries first compress information into the corresponding registers; registers from different frames then exchange information through self-attention; and the aggregated context is propagated back to the point queries. This compress--communicate--broadcast process enables scene-wide information exchange while shifting the expensive interaction from point space to a substantially smaller register space.

Dynamic instances require a different interaction structure. Following PointForward and related dynamic reconstruction methods~\cite{yang2023unisim, yan2024street, chen2025omnire, chi2026pointforward}, we align each moving instance's observations in an object-centric canonical space and assign each instance its own Dynamic Instance Registers. This enables \emph{intra-instance interaction} among canonicalized observations while keeping different instances independent. The resulting topology follows the composition of driving scenes: static observations interact across frames, whereas dynamic observations interact within their respective canonical spaces. We further employ 4D rotary positional encoding~\cite{su2024roformer} using aligned spatial coordinates and original timestamps, preserving temporal provenance during register-mediated interaction.

We evaluate PointForward V2 on three autonomous-driving benchmarks. On Waymo~\cite{sun2020scalability}, it improves full-image PSNR over PointForward from 28.48 to 29.88 and reduces dynamic-region D-RMSE from 6.07 to 3.99. The improvements remain consistent on nuScenes~\cite{caesar2020nuscenes} and Argoverse 2~\cite{wilson2023argoverse} under both zero-shot and fine-tuning settings. Comprehensive ablations further validate the effectiveness of point interaction and scene-structured registers.

Our contributions can be summarized as follows:

\begin{itemize}
    \item We propose \textbf{PointForward V2}, advancing feedforward driving reconstruction from point-aligned representation to point-aligned interaction and enabling efficient spatio-temporal information exchange among dense 3D queries.
    \item We introduce scene-structured register interaction, where Static Scene Registers enable inter-frame scene interaction for the static background. Moreover, Dynamic Instance Registers enable intra-instance interaction among canonicalized observations of each moving instance.
    \item Extensive experiments on Waymo, nuScenes, and Argoverse 2 show consistent improvements over previous feedforward reconstruction methods, with comprehensive ablations validating the proposed interaction framework.
\end{itemize}
\section{Related Work}
\label{sec:work}

\begin{figure*}[t]
    \centering
    \includegraphics[width=1.0\textwidth]{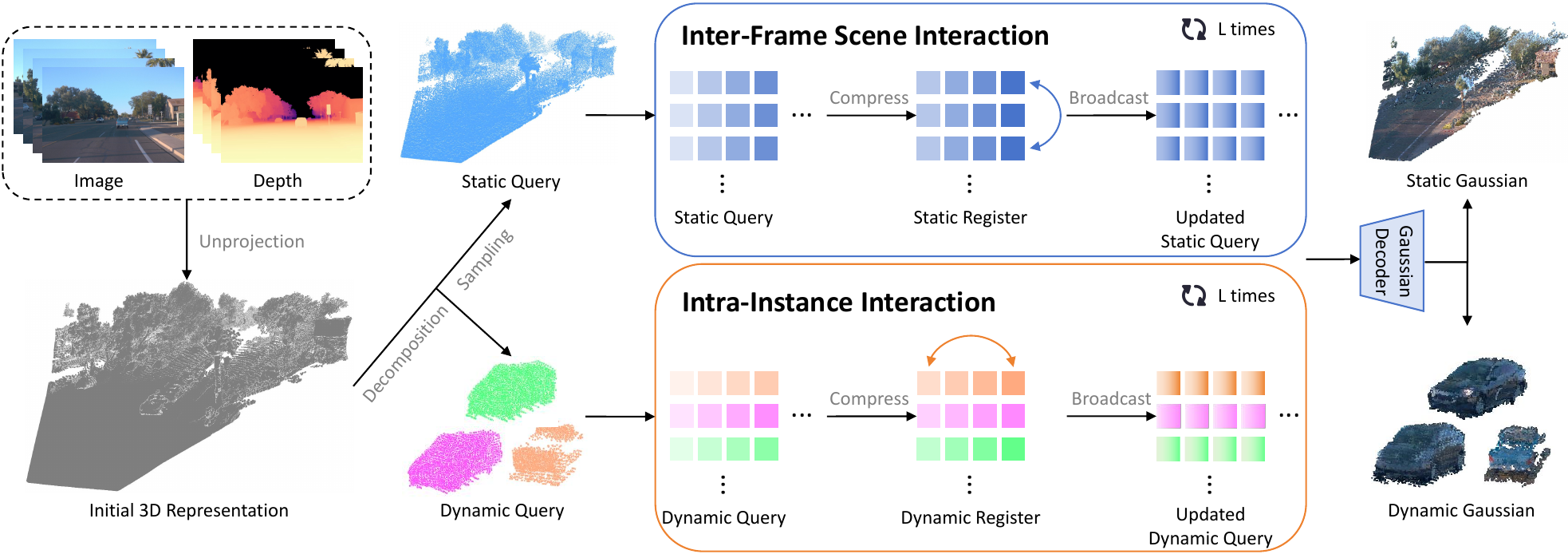}
    \caption{\textbf{Overview of PointForward V2.} It first unprojects images into a 3D representation and decomposes it into static and dynamic point queries. Inter-Frame Scene Interaction refines static queries, while Intra-Instance Interaction refines dynamic queries. Both branches use a compress--communicate--broadcast register mechanism repeated for $L$ blocks. Updated queries are finally decoded into 3D Gaussians.}
    \label{fig:overview}
    \vspace{-10px}
\end{figure*}

\textbf{Feedforward Driving Scene Reconstruction.} Feedforward reconstruction methods learn cross-scene priors to predict explicit 3D representations directly without per-scene optimization. Early approaches such as pixelSplat and MVSplat~\cite{charatan2024pixelsplat, chen2024mvsplat} predict pixel-aligned 3D Gaussians from posed images, while later methods extend this paradigm to uncalibrated or pose-free inputs~\cite{smart2024splatt3r, chen2024pref3r, jiang2025anysplat}. Recent works further address dynamic driving scenes. DrivingForward~\cite{tian2025drivingforward} targets real-time reconstruction from surround-view inputs, while STORM~\cite{yang2025storm} models scene dynamics through Gaussian motion. DGGT~\cite{chen2026dggt} jointly estimates camera parameters, Gaussian representations, and scene dynamics for pose-free 4D reconstruction, while ReconDrive~\cite{yu2026recondrive} leverages geometric foundation models for static--dynamic reconstruction. Instant NuRec~\cite{huang2026instant} reconstructs driving logs into static and dynamic Gaussian representations in a single forward pass. PointForward~\cite{chi2026pointforward} introduces sparse 3D queries for point-aligned feedforward driving reconstruction; PointForward V2 builds on this representation and focuses on efficient interaction.

\textbf{Structured Dynamic Scene Representation.} Dynamic driving scenes naturally consist of a largely static background and multiple moving instances. Neural Scene Graphs~\cite{ost2021neural}, UniSim~\cite{yang2023unisim}, and SUDS~\cite{turki2023suds} exploit explicit scene structure and static--dynamic decomposition for dynamic-scene modeling, while EmerNeRF~\cite{yang2024emernerf} models scene dynamics with static and dynamic neural fields and scene flow. Recent Gaussian-based methods, including Street Gaussians~\cite{yan2024street}, DeSiRe-GS~\cite{peng2025desire}, and OmniRe~\cite{chen2025omnire}, further represent object-centric dynamics with explicit Gaussians. PointForward~\cite{chi2026pointforward} organizes dynamic instances in object-centric canonical coordinates while maintaining the static background in a shared world coordinate system. PointForward V2 further uses this compositional structure to define the interaction scope: static observations interact across frames, whereas dynamic observations interact only within their corresponding instances.

\textbf{Efficient Global Interaction.} Global self-attention enables long-range information exchange but scales quadratically. Prior work reduces this cost using compact intermediate representations, including inducing points in Set Transformer~\cite{lee2019set}, latent arrays in Perceiver~\cite{jaegle2021perceiver} and Perceiver IO~\cite{jaegle2021perceiverio}, and adaptive or proxy tokens in TokenLearner~\cite{ryoo2021tokenlearner} and Agent Attention~\cite{han2024agent}. Similar ideas have also been explored in geometric reasoning: CoTracker~\cite{karaev2024cotracker} uses token proxies for large sets of point tracks, while DELTA~\cite{ngo2025delta} adopts efficient global--local attention for long-range 3D tracking. Most closely related, VGGT-$\Omega$~\cite{wang2026vggt} restricts inter-frame communication to compact register tokens. PointForward V2 follows this compact-mediated interaction principle, but organizes the communication bottlenecks according to driving-scene composition: Static Scene Registers enable inter-frame interaction, while Dynamic Instance Registers restrict interaction to each canonicalized moving instance.
\section{Method}
\label{sec:method}

Given $T$ multi-view context images, PointForward V2 reconstructs the dynamic scene in a single forward pass. As shown in Fig.~\ref{fig:overview}, we first lift image observations into sparse point-aligned queries and decompose them into the static background and dynamic instances. Static queries remain in the shared world coordinate system, while dynamic queries align in object-centric canonical spaces. The core of PointForward V2 is \emph{scene-structured register interaction}: static queries exchange information across views through Static Scene Registers, whereas dynamic queries interact independently within each instance through Dynamic Instance Registers. Unlike PointForward~\citep{chi2026pointforward}, which independently aggregates cross-view evidence for each query through image-space reprojection, PointForward V2 directly enables information exchange among lifted point representations through compact registers.

\subsection{Point-Aligned Scene Representation}

We build upon the point-aligned representation of PointForward~\cite{chi2026pointforward}. Each lifted observation is associated with a 3D position $\mathbf{x} \in \mathbb{R}^3$, RGB value $\mathbf{c} \in \mathbb{R}^3$, image feature $\mathbf{f} \in \mathbb{R}^{C_f}$, Plücker ray $\mathbf{r} \in \mathbb{R}^6$, and timestamp $\tau \in \mathbb{R}$. Static observations remain in the shared world coordinate system. For a dynamic observation with world-space position $\mathbf{x}^w \in \mathbb{R}^3$ and object-to-world pose $\mathbf{T} \in SE(3)$, we transform its position into the object-centric canonical space:
\begin{equation}
    \mathbf{x}^{\mathrm{can}}=\mathbf{T}^{-1}\mathbf{x}^{w}.
\end{equation}
Only the spatial coordinate is canonicalized, while the original timestamp and Plücker ray are retained. This design aligns observations of the same moving object geometrically across frames without discarding their observation provenance. In particular, canonicalized coordinates provide a common object-relative reference for interaction, whereas the original timestamp and viewing ray preserve when and from where each observation was acquired.

We sample $N_s$ static observations per context view and $N_d$ dynamic observations per instance. Each sampled observation is initialized as a query $\mathbf{q} \in \mathbb{R}^C$:
\begin{equation}
\mathbf{q} = \phi_{\mathrm{init}}\left(\mathbf{x},\mathbf{r},\mathbf{c},\mathbf{f}\right),
\end{equation}
where $\phi_{\mathrm{init}}: \mathbb{R}^{12+C_f} \rightarrow \mathbb{R}^C$ embeds and fuses the input features. We denote the static queries from view $t$ as $\mathbf{Q}_t^s \in \mathbb{R}^{N_s \times C}$ and the dynamic queries of instance $k$ as $\mathbf{Q}_k^d \in \mathbb{R}^{N_d \times C}$. The compress and broadcast stages already induce information exchange among point queries: each register aggregates information from multiple points, and each point subsequently reads from these shared register representations.

\subsection{Scene-Structured Register Interaction}

Direct global self-attention among all point queries incurs quadratic complexity, making it prohibitively expensive for large driving scenes. We therefore introduce compact learnable registers as communication bottlenecks between point queries. Specifically, we maintain shared register banks $\mathbf{R}^s \in \mathbb{R}^{R_s\times C}$ and $\mathbf{R}^d \in \mathbb{R}^{R_d \times C}$, which are instantiated for each context view and dynamic instance, respectively. Rather than allowing all point queries to attend to one another directly, each interaction block follows a \emph{compress--communicate--broadcast} procedure: registers first aggregate information from dense point queries, interaction is then performed in the compact register space, and the resulting context is finally propagated back to the point queries.

We denote by $\mathcal{A}(\mathbf{Q}, \mathbf{C})$ a cross-attention block, where $\mathbf{Q} \in \mathbb{R}^{N_q \times C}$ provides the attention queries and $\mathbf{C} \in \mathbb{R}^{N_c \times C}$ provides the keys and values, producing an output in $\mathbb{R}^{N_q \times C}$. We denote self-attention by $\mathcal{S}(\cdot)$. For clarity, we omit the interaction-block index below.

\textbf{Inter-Frame Scene Interaction.} Static observations from different context views describe the same background in a shared world coordinate system, and therefore benefit from scene-level information exchange across frames. For each view $t$, we instantiate $\mathbf R_t^s\in\mathbb R^{R_s\times C}$. The registers first attend to the static queries of the corresponding view, compressing the dense point representation into a compact scene representation:
\begin{equation}
    \begin{aligned}
        \hat{\mathbf{R}}_t^s &= \mathcal{A}\left(\mathbf{R}_t^s,\mathbf{Q}_t^s\right), \\
        [\tilde{\mathbf{R}}_1^s; \ldots; \tilde{\mathbf{R}}_T^s] &= \mathcal{S}\left([\hat{\mathbf{R}}_1^s; \ldots; \hat{\mathbf{R}}_T^s]\right), \\
        \tilde{\mathbf {Q}}_t^s &= \mathcal{A}\left(\mathbf{Q}_t^s,\tilde{\mathbf{R}}_t^s\right),
    \end{aligned}
\end{equation}
where $\hat{\mathbf{R}}_t^s, \tilde{\mathbf{R}}_t^s \in \mathbb{R}^{R_s \times C}$ and $\tilde{\mathbf{Q}}_t^s \in \mathbb{R}^{N_s \times C}$. Thus, the pathway $\mathbf Q^s_t \rightarrow \mathbf R^s_t \rightarrow \mathbf Q^s_t$ acts as a compact-mediated interaction among dense point queries. Register self-attention further extends this interaction across context views by allowing their compressed scene representations to communicate before being broadcast back to the point space. Importantly, point queries from different views never directly attend to one another; all cross-frame communication is mediated by the compact Static Scene Registers.

\textbf{Intra-Instance Interaction.} Dynamic observations require a different interaction scope. Queries belonging to the same moving instance are first spatially aligned in its object-centric canonical space, so observations collected at different timestamps share a common object-relative coordinate system. For each instance $k$, we instantiate $\mathbf R_k^d\in\mathbb R^{R_d\times C}$ and perform:
\begin{equation}
    \begin{aligned}
        \hat{\mathbf{R}}_k^d &= \mathcal{A}\left(\mathbf{R}_k^d,\mathbf{Q}_k^d\right), \\
        \tilde{\mathbf{R}}_k^d &= \mathcal{S}\left(\hat{\mathbf{R}}_k^d\right), \\
        \tilde{\mathbf {Q}}_k^d &= \mathcal{A}\left(\mathbf{Q}_k^d,\tilde{\mathbf{R}}_k^d\right),
    \end{aligned}
\end{equation}
where $\hat{\mathbf{R}}_k^d, \tilde{\mathbf{R}}_k^d \in \mathbb{R}^{R_d \times C}$ and $\tilde{\mathbf{Q}}_k^d \in \mathbb{R}^{N_d \times C}$. Each instance performs independent register interaction, preventing information exchange between unrelated dynamic objects. Because $\mathbf Q_k^d$ aggregates canonicalized observations of the same instance across context views, its registers summarize temporally distributed observations in a common object-relative coordinate system. Restricting communication within each instance avoids mixing unrelated object motions while allowing complementary observations of the same object to refine its representation.

\textbf{Complexity.} We stack $L$ interaction blocks and propagate both query and register features across blocks. Applying global self-attention to the $TN_s$ static queries requires $\mathcal{O}((TN_s)^2)$ computation, whereas our static interaction requires $\mathcal{O}(TN_sR_s + (TR_s)^2)$. For $K$ dynamic instances, the corresponding complexity is $\mathcal{O}(KN_dR_d + KR_d^2)$.

\subsection{Aligned 4D Positional Encoding}
\label{sec:encoding}

Registers themselves are position-agnostic, while point queries retain explicit geometric and temporal structure. We therefore associate each point query with an aligned 4D coordinate $\mathbf p=[\mathbf x;\tau]\in\mathbb R^4$. The spatial component $\mathbf x$ ensures that attention respects the coordinate system in which interaction is performed: world coordinates for static queries and canonical object coordinates for dynamic queries. The timestamp $\tau$ remains in the original temporal coordinate, allowing spatially aligned observations of the same dynamic instance to remain temporally distinguishable.

\begin{table*}[t]
    \centering
        \begin{tabular}{lcccccc}
            \hline
            \multirow{2}{*}{Method} & \multicolumn{3}{c}{Dynamic Only} & \multicolumn{3}{c}{Full Image} \\
             & PSNR↑ & SSIM↑ & D-RMSE↓ & PSNR↑ & SSIM↑ & D-RMSE↓ \\ \hline
            \multicolumn{7}{l}{\emph{Per-Scene Optimization Methods}} \\
            EmerNeRF~\citep{yang2024emernerf} & 17.79 & 0.255 & 40.88 & 24.51 & 0.738 & 33.99 \\
            3DGS~\citep{kerbl20233d} & 17.13 & 0.267 & 13.88 & 25.13 & 0.741 & 19.68 \\
            PVG~\citep{chen2026periodic} & 15.51 & 0.128 & 15.91 & 22.38 & 0.661 & 13.01 \\
            DeformableGS~\citep{yang2024deformable} & 17.10 & 0.266 & 12.14 & 25.29 & 0.761 & 14.79 \\ \hline
            \multicolumn{7}{l}{\emph{Generalizable Feedforward Methods}} \\
            LGM~\citep{tang2024lgm} & 19.58 & 0.443 & 9.43 & 23.59 & 0.691 & 8.02 \\
            GS-LRM~\citep{zhang2024gs} & 20.02 & 0.520 & 9.95 & 25.18 & 0.753 & 7.94 \\
            Depth Anything 3~\citep{lin2025depth} & 15.96 & 0.429 & 12.20 & 23.27 & 0.686 & 11.66 \\
            AnySplat~\citep{jiang2025anysplat} & 15.99 & 0.418 & 10.45 & 23.32 & 0.687 & 10.31 \\
            STORM~\citep{yang2025storm} & 22.10 & 0.624 & 7.50 & 26.38 & 0.794 & 5.48 \\
            DGGT~\citep{chen2026dggt} & 22.80 & 0.651 & 6.55 & 27.41 & 0.846 & 3.47 \\
            PointForward~\citep{chi2026pointforward} & 25.01 & 0.768 & 6.07 & 28.48 & 0.861 & 3.35 \\
            PointForward V2 (Ours) & \textbf{25.81} & \textbf{0.773} & \textbf{3.99} & \textbf{29.88} & \textbf{0.865} & \textbf{2.89} \\ \hline
        \end{tabular}
    \vspace{-5px}
    \caption{\textbf{Quantitative results on Waymo.} We report PSNR, SSIM, and D-RMSE over both dynamic regions and full images. PointForward V2 achieves the best performance across all metrics, with particularly great improvements on dynamic regions.}
    \label{tab:waymo}
    \vspace{-5px}
\end{table*}

We encode $\mathbf{p}$ using a 4D extension of rotary positional encoding (RoPE)~\cite{su2024roformer}, with the four coordinates assigned to separate channel groups within each attention head. Let $\mathcal{R}_{\mathbf{p}}(\cdot)$ denote the corresponding shape-preserving rotary transformation and $\mathbf{W}_Q,\mathbf{W}_K,\mathbf{W}_V \in \mathbb{R}^{C \times C}$ the attention projections. RoPE is applied only to the point side of query--register cross-attention:

\begin{equation}
    \begin{aligned}
        \text{Compress:}\quad
        &\mathbf Q_{\mathrm{attn}}=\mathbf R\mathbf W_Q,\\
        &\mathbf K_{\mathrm{attn}}=\mathcal R_{\mathbf p}(\mathbf Q\mathbf W_K),\quad
        \mathbf V_{\mathrm{attn}}=\mathbf Q\mathbf W_V,\\
        \text{Broadcast:}\quad
        &\mathbf Q_{\mathrm{attn}}=\mathcal R_{\mathbf p}(\mathbf Q\mathbf W_Q),\\
        &\mathbf K_{\mathrm{attn}}=\mathbf R\mathbf W_K,\quad
        \mathbf V_{\mathrm{attn}}=\mathbf R\mathbf W_V.
    \end{aligned}
\end{equation}

During register self-attention, no positional encoding is applied because registers do not correspond to explicit geometric locations. Together with the original Plücker ray retained in each point query, this design separates three complementary cues: aligned coordinates encode interaction geometry, timestamps preserve temporal provenance, and Plücker rays preserve the original viewing provenance.

\subsection{Gaussian Prediction and Rendering}

After $L$ interaction blocks, a shared Gaussian decoder $\phi_{\mathrm{gs}}:\mathbb{R}^C \rightarrow \mathbb{R}^{M \times 15}$ decodes each updated query $\mathbf{q} \in \mathbb{R}^C$ into $M$ 3D Gaussians. The $m$-th Gaussian $\mathbf{g}_m \in \mathbb{R}^{15}$ is parameterized as:
\begin{equation}
    \mathbf{g}_m = \phi_{\mathrm{gs}}(\mathbf{q})_m = \left(\Delta \mathbf{x}_m, \Delta \mathbf{c}_m, \mathbf{s}_m, \boldsymbol{\rho}_m, o_m, \sigma_m\right),
\end{equation}
where $\Delta \mathbf{x}_m, \Delta \mathbf{c}_m, \mathbf{s}_m \in \mathbb{R}^3$, $\boldsymbol{\rho}_m \in \mathbb{R}^4$, and $o_m, \sigma_m \in \mathbb{R}$ denote position residual, color residual, scale, rotation, opacity, and temporal lifespan, respectively. We predict position and color as residuals with respect to the sampled observation, anchoring each Gaussian to the underlying point-aligned input while allowing the decoder to refine its geometry and appearance. Predicting multiple Gaussians per query further increases local representation capacity without increasing the number of point queries, enabling each interacted point representation to describe richer nearby scene content.

Static Gaussians are directly represented in the world coordinate system. Dynamic Gaussians are predicted in their object-centric canonical spaces and transformed to the world space using the instance pose at the target timestamp. Following PointForward~\cite{chi2026pointforward}, the lifespan modulates opacity according to the temporal distance between the source observation and the target timestamp.

\subsection{Training Objectives}

We train PointForward V2 using image reconstruction, depth, and perceptual supervision. The overall objective $\mathcal{L} \in \mathbb{R}$ is:
\begin{equation}
    \mathcal{L} = \mathcal{L}_{\mathrm{rgb}} + \lambda_{\mathrm{depth}} \mathcal{L}_{\mathrm{depth}} + \lambda_{\mathrm{lpips}} \mathcal{L}_{\mathrm{lpips}},
\end{equation}
where $\mathcal{L}_{\mathrm{rgb}}, \mathcal{L}_{\mathrm{depth}}, \mathcal{L}_{\mathrm{lpips}} \in \mathbb{R}$ denote the RGB reconstruction, depth, and perceptual losses, respectively, and $\lambda_{\mathrm{depth}}, \lambda_{\mathrm{lpips}}\in\mathbb{R}^{+}$ are their weights.
\section{Experiment}
\label{sec:exp}

\subsection{Implementation Details}

We train PointForward V2 on 8 NVIDIA H20 GPUs with a global batch size of 8 using AdamW and a cosine learning-rate schedule. Following PointForward~\cite{chi2026pointforward}, we sample 20-frame clips from Waymo's three frontal cameras, using 4 frames as context views. We resize input images to $512 \times 768$, use ImageNet-pretrained EfficientNetV2-S~\cite{tan2021efficientnetv2} as the image encoder, and obtain input depth maps following PointForward's settings. We set $N_s=65{,}536$, $N_d=4{,}096$, $R_s=256$, $R_d=16$, $L=4$, $M=4$, $\lambda_{depth}=1$, and $\lambda_{lpips}=0.05$.

\begin{figure*}[t]
    \centering
    \includegraphics[width=1.0\textwidth]{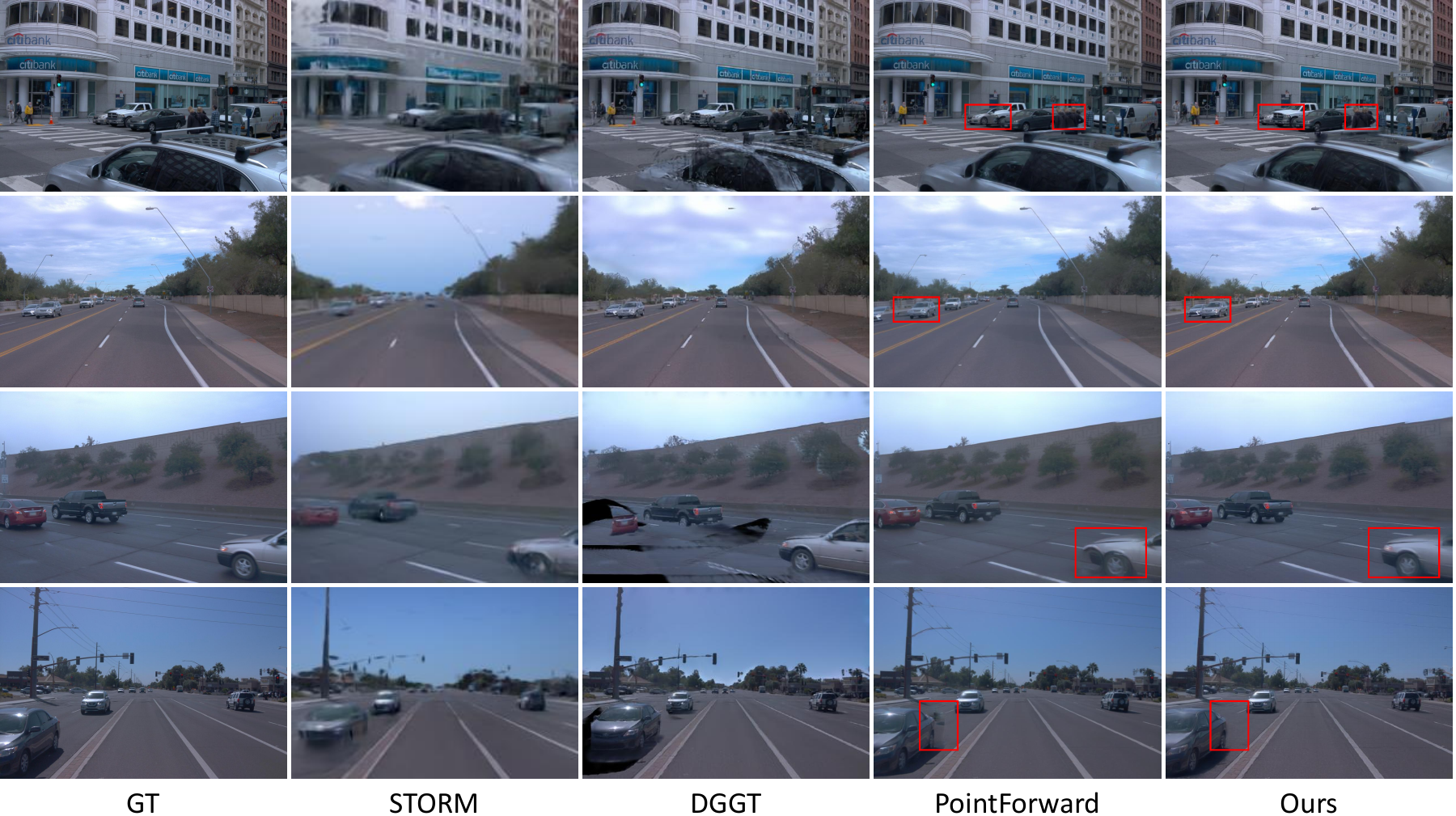}
    \caption{\textbf{Qualitative comparison on Waymo.} Compared with STORM, DGGT, and PointForward, PointForward V2 produces more complete scene geometry and cleaner reconstructions of both static and dynamic content, with improved object shapes and scene structures.}
    \label{fig:waymo}
    \vspace{-10px}
\end{figure*}

\subsection{Datasets and Metrics}

We evaluate our model on Waymo Open Dataset~\cite{sun2020scalability}, nuScenes~\cite{caesar2020nuscenes}, and Argoverse 2~\cite{wilson2023argoverse}. For Waymo, we follow PointForward's training and evaluation protocol~\cite{chi2026pointforward}. We report PSNR, SSIM, and D-RMSE for novel-view reconstruction. For a fair comparison across methods with different rendering resolutions, we resize our predictions to a common resolution before computing metrics, using $384 \times 576$ for Waymo. The appendix provides the evaluation protocol and resolutions for other datasets.

\subsection{Comparison with State-of-the-Art Methods}

\textbf{Baselines.} We compare PointForward V2 with both per-scene optimization and generalizable feedforward reconstruction methods. On Waymo, we include EmerNeRF~\cite{yang2024emernerf}, 3DGS~\cite{kerbl20233d}, PVG~\cite{chen2026periodic}, and DeformableGS~\cite{yang2024deformable} as optimization-based baselines, together with recent feedforward approaches including LGM~\cite{tang2024lgm}, GS-LRM~\cite{zhang2024gs}, Depth Anything 3~\cite{lin2025depth}, AnySplat~\cite{jiang2025anysplat}, STORM~\cite{yang2025storm}, DGGT~\cite{chen2026dggt}, and PointForward~\cite{chi2026pointforward}. For DGGT, we follow PointForward's evaluation setting, which disables its diffusion-based rendering refinement for qualitative visualization. This module is not specific to DGGT and can be applied to any method to refine rendered images. Removing it allows for a direct comparison of the raw Gaussian rendering quality across methods.

\textbf{Waymo.} As shown in Tab.~\ref{tab:waymo}, PointForward V2 consistently outperforms previous methods on both full images and dynamic regions. Compared with PointForward, our method improves full-image PSNR from 28.48 to 29.88 and reduces D-RMSE from 3.35 to 2.89. The improvement is particularly pronounced in dynamic regions, where D-RMSE decreases from 6.07 to 3.99, demonstrating the benefit of point-aligned interaction. These gains indicate that information exchange among aligned point queries improves not only appearance reconstruction but also the geometric consistency of the recovered scene. Qualitative results in Fig.~\ref{fig:waymo} further show more complete scene structures and more coherent dynamic-object reconstruction compared with STORM, DGGT, and PointForward.

\begin{figure*}
    \centering
    \includegraphics[width=1.0\textwidth]{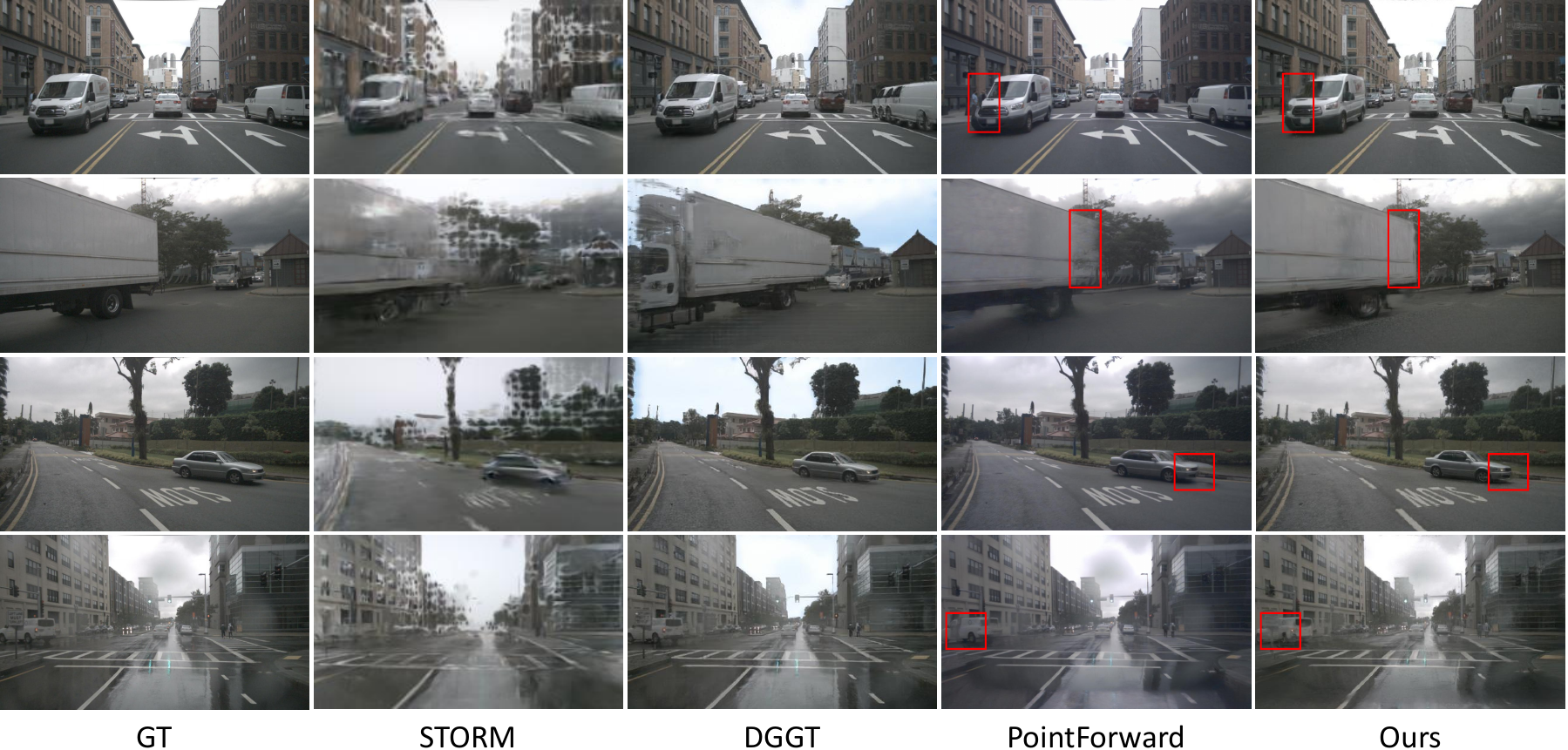}
    \caption{\textbf{Zero-shot qualitative comparison on nuScenes.} Models are evaluated without nuScenes-specific fine-tuning. PointForward V2 better preserves scene structures and dynamic objects with cleaner geometry and more complete reconstructions.}
    \label{fig:nuscenes}
    \vspace{-10px}
\end{figure*}

\textbf{Additional Datasets.} We further evaluate PointForward V2 on nuScenes and Argoverse 2 under both zero-shot and fine-tuning settings. As shown in Tab.~\ref{tab:dataset}, our method consistently improves over STORM, DGGT, and PointForward on both datasets. Under zero-shot evaluation, PointForward V2 improves PSNR over PointForward by 1.83 dB on nuScenes and 0.99 dB on Argoverse 2. The gains remain consistent after fine-tuning, reaching 29.22 PSNR on nuScenes and 28.63 PSNR on Argoverse 2. The zero-shot comparisons in Fig.~\ref{fig:nuscenes} further demonstrate improved reconstruction of scene geometry and dynamic content under a different data distribution.

\begin{table*}[t]
    \centering

    \begin{minipage}[t]{0.49\textwidth}
        \vspace{0pt}
        \centering
        \begin{tabular}{lcccc}
            \hline
            \multirow{2}{*}{Method}
            & \multicolumn{2}{c}{nuScenes}
            & \multicolumn{2}{c}{Argoverse 2} \\
            & PSNR$\uparrow$ & SSIM$\uparrow$
            & PSNR$\uparrow$ & SSIM$\uparrow$ \\ \hline
            \multicolumn{5}{l}{\emph{Zero-Shot}} \\
            STORM & 17.77 & 0.669 & 20.83 & 0.542 \\
            DGGT & 25.31 & 0.794 & 25.02 & 0.801 \\
            PointForward & 26.54 & 0.821 & 25.50 & 0.810 \\
            Ours & \textbf{28.37} & \textbf{0.829}
                 & \textbf{26.49} & \textbf{0.824} \\ \hline
            \multicolumn{5}{l}{\emph{Fine-Tuning}} \\
            STORM & 24.54 & 0.784 & 24.97 & 0.791 \\
            DGGT & 26.63 & 0.813 & 26.96 & 0.831 \\
            PointForward & 27.50 & 0.826 & 27.66 & 0.851 \\
            Ours & \textbf{29.22} & \textbf{0.847}
                 & \textbf{28.63} & \textbf{0.872} \\ \hline
        \end{tabular}

        \captionsetup{skip=3pt}
        \caption{\textbf{Additional datasets.}
        Quantitative results on nuScenes and Argoverse 2 under zero-shot
        and fine-tuning settings.}
        \label{tab:dataset}
    \end{minipage}
    \hfill
    \begin{minipage}[t]{0.49\textwidth}
        \vspace{0pt}
        \centering

        \begin{tabular}{lccc}
            \hline
            Variant & PSNR$\uparrow$ & SSIM$\uparrow$ & D-RMSE$\downarrow$ \\ \hline
            w/o All Attention & 21.42 & 0.653 & 7.56 \\
            w/o Reg. SA & 28.78 & 0.848 & 2.90 \\
            w/o 4D RoPE & 29.58 & 0.858 & 2.90 \\
            Ours & \textbf{29.88} & \textbf{0.865} & \textbf{2.89} \\ \hline
        \end{tabular}

        \captionsetup{skip=3pt}
        \caption{\textbf{Interaction ablation on Waymo.}}
        \label{tab:ablation}

        \vspace{0.8em}

        \begin{tabular}{cccc}
            \hline
            Gaussian & PSNR$\uparrow$ & SSIM$\uparrow$ & D-RMSE$\downarrow$ \\ \hline
            1 & 28.62 & 0.841 & 2.95 \\
            2 & 29.46 & 0.858 & \textbf{2.88} \\
            4 & \textbf{29.88} & \textbf{0.865} & 2.89 \\
            8 & 29.67 & 0.862 & 2.98 \\ \hline
        \end{tabular}

        \captionsetup{skip=3pt}
        \caption{\textbf{Gaussians per query.} We study the effect of predicting different numbers of Gaussians from each point query.}
        \label{tab:gaussian}
    \end{minipage}
\end{table*}

\begin{figure*}
  \centering
  \includegraphics[width=1.0\textwidth]{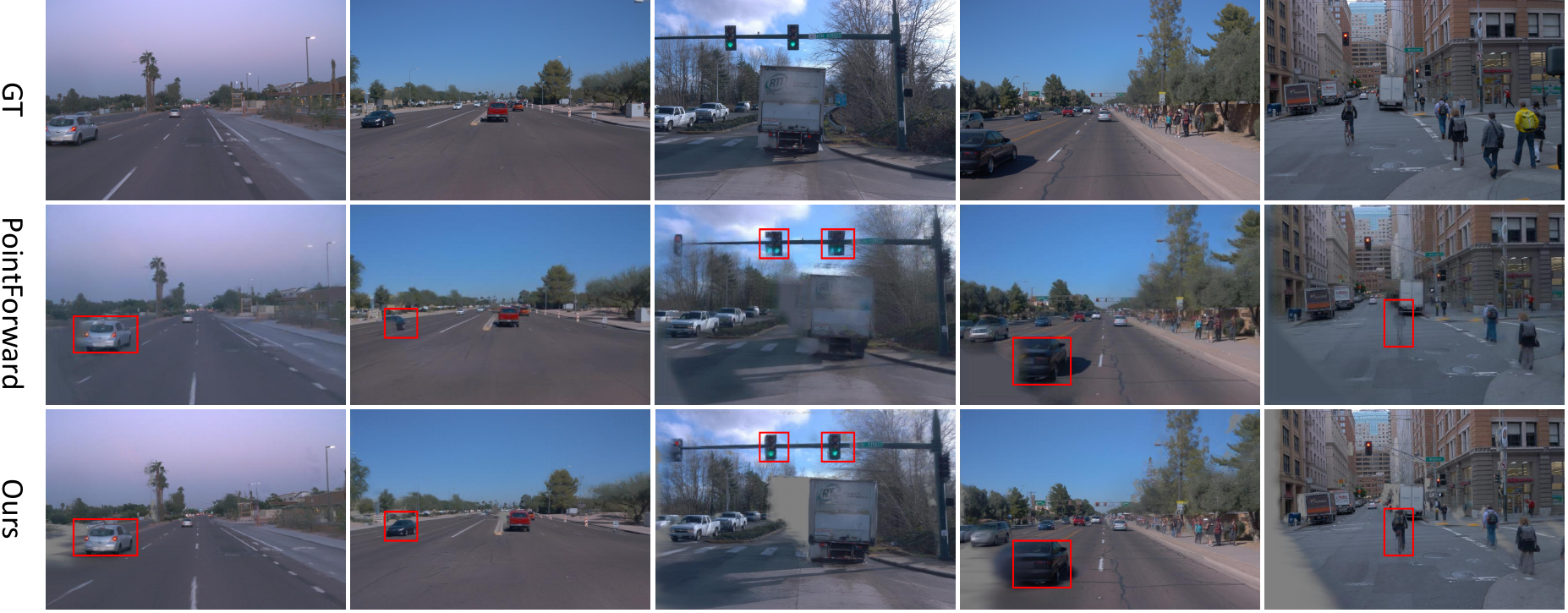}
  \caption{\textbf{Qualitative comparison under extrapolated viewpoints.} PointForward renders feature maps followed by convolutional RGB refinement, which can fill newly disoccluded regions but may also introduce incorrect image-space completion. PointForward V2 directly renders RGB Gaussians, avoiding such erroneous completion and preserving the reconstructed 3D content more faithfully.}
  \label{fig:novel}
\end{figure*}

\textbf{Extrapolated View.} Following PointForward~\cite{chi2026pointforward}, we additionally examine rendering from extrapolated viewpoints beyond the observed camera trajectory. Fig.~\ref{fig:novel} highlights an important difference between the rendering designs of PointForward and PointForward V2. PointForward renders learned feature maps and subsequently converts them to RGB using convolutional refinement. Such spatial processing can plausibly fill regions that become newly visible under viewpoint changes. However, this completion is not explicitly constrained by the reconstructed 3D geometry and can therefore introduce incorrect structures. In contrast, PointForward V2 directly renders RGB-valued Gaussians. As shown in Fig.~\ref{fig:novel}, this avoids several erroneous completions produced by PointForward and preserves the underlying reconstructed content more faithfully.

\subsection{Ablation Studies}

\textbf{Component Ablation.} We first evaluate the contribution of the proposed interaction components in Tab.~\ref{tab:ablation}. Removing all attention leads to a large performance drop, reducing PSNR from 29.88 to 21.42, confirming that interaction among point queries is critical for accurate reconstruction. Notably, retaining only the query--register cross-attention already recovers most of the performance, reaching 28.78 PSNR even without register self-attention. This shows that the compress--broadcast pathway itself enables effective indirect query interaction: the registers first aggregate information from multiple point queries and then redistribute the shared context back to them, forming a compact $\mathbf{Q}\!\rightarrow\!\mathbf{R}\!\rightarrow\!\mathbf{Q}$ communication path. Register self-attention further improves the result to 29.88 PSNR by using the compressed representations to exchange information before being broadcast back to the dense point queries. This enables scene-level communication across different context views. Removing 4D RoPE causes only a small but consistent degradation, suggesting that aligned spatial and temporal cues provide complementary positional information for register-mediated interaction.

\begin{table}
    \centering
    \begin{tabular}{lccc}
        \hline
        Capacity & PSNR↑ & SSIM↑ & D-RMSE↓ \\ \hline
        \multicolumn{4}{l}{\emph{$R_s / R_d$}} \\
        128 / 8 & 29.31 & 0.856 & 2.96 \\
        256 / 16 & \textbf{29.88} & \textbf{0.865} & \textbf{2.89} \\
        512 / 32 & 29.68 & 0.861 & 2.90 \\ \hline
        \multicolumn{4}{l}{\emph{$N_s / N_d$}} \\
        16,384 / 1,024 & 28.61 & 0.843 & 2.94 \\
        32,768 / 2,048 & 29.21 & 0.854 & 2.91 \\
        65,536 / 4,096 & \textbf{29.88} & \textbf{0.865} & \textbf{2.89} \\ \hline
    \end{tabular}
    \vspace{-5px}
    \caption{\textbf{Effect of register and query capacity.} We vary register sizes $R_s / R_d$ and query numbers $N_s / N_d$ independently while keeping the remaining configuration fixed.}
    \label{tab:capacity}
    \vspace{-5px}
\end{table}

\textbf{Gaussians per Query.} We further study the number of Gaussians $M$ decoded from each point query in Tab.~\ref{tab:gaussian}. Increasing $M$ from 1 to 4 steadily improves PSNR and SSIM, indicating that multiple Gaussians provide greater local representation capacity around each sampled point. Increasing $M$ further to 8 yields no additional benefit and slightly degrades all three metrics. This suggests that a small set of Gaussians is sufficient to capture the local structure around each query. We therefore use $M=4$ by default, which achieves the best PSNR and SSIM.

\textbf{Register and Query Capacity.} Tab.~\ref{tab:capacity} studies the capacity of the point and register representations. With the number of point queries fixed, increasing $R_s/R_d$ from $128/8$ to $256/16$ improves reconstruction, while increasing it to $512/32$ brings no additional benefit. This suggests that a compact register bottleneck is sufficient for mediating point interaction. In contrast, increasing the number of point queries consistently improves reconstruction quality. These results show that PointForward V2 can benefit from increasingly dense point representations without requiring a proportional increase in interaction tokens.
\section{Conclusion}
\label{sec:con}

We presented PointForward V2, which advances feedforward driving reconstruction from point-aligned representation to point-aligned interaction. Building upon the sparse 3D queries, we introduce scene-structured registers that enable efficient inter-frame interaction for static scenes and intra-instance interaction for dynamic objects. Together with aligned 4D positional encoding, PointForward V2 supports scalable spatio-temporal reasoning without dense global attention. Experiments on Waymo, nuScenes, and Argoverse 2 demonstrate state-of-the-art reconstruction performance, while extensive ablations validate the effectiveness and scalability of the proposed interaction framework.

\textbf{Limitations.} PointForward V2 currently uses fixed query and register budgets across scenes and dynamic instances. In addition, our experiments focus on large-scale autonomous driving datasets; extending point-aligned interaction to more diverse dynamic environments and reconstruction settings remains an interesting direction.

{
    \small
    \bibliographystyle{ieeenat_fullname}
    \bibliography{main}
}

% WARNING: do not forget to delete the supplementary pages from your submission 
% \input{sec/X_suppl}

\end{document}